\documentclass[10pt,twocolumn,letterpaper]{article}

\usepackage[pagenumbers]{cvpr} 

\usepackage{graphicx}
\usepackage{amsmath}
\usepackage{amssymb}
\usepackage{booktabs}
\usepackage{vcell}
\usepackage{dblfloatfix}
\usepackage[pagebackref,breaklinks,colorlinks]{hyperref}

\usepackage[capitalize]{cleveref}
\crefname{section}{Sec.}{Secs.}
\Crefname{section}{Section}{Sections}
\Crefname{table}{Table}{Tables}
\crefname{table}{Tab.}{Tabs.}

\def\cvprPaperID{7} 
\def\confName{CVPR}
\def\confYear{2022}

\newcommand*{\affaddr}[1]{#1} 
\newcommand*{\affmark}[1][*]{\textsuperscript{#1}}
\newcommand*{\email}[1]{\texttt{#1}}

\begin{document}

\title{Empirical Advocacy of Bio-inspired Models for Robust Image Recognition}




\author{%
Harshitha Machiraju\affmark[1,2]$^\star$, Oh-Hyeon Choung\affmark[1]$^\star$, Michael H. Herzog\affmark[1], and Pascal Frossard\affmark[2] \\
\small{\affaddr{\affmark[1]Laboratory of Psychophysics (LPSY)},\affaddr{\affmark[2]Signal Processing Laboratory 4 (LTS4)}}\\
\small{\affaddr{Ecole Polytechnique Fédérale de Lausanne (EPFL), Switzerland}}\\
\small{\email{\{harshitha.machiraju,michael.herzog,pascal.frossard\}@epfl.ch}, \email{ohhyeon.choung@gmail.com}}
}

\maketitle
\begin{abstract}
Deep convolutional neural networks (DCNNs) have revolutionized computer vision and are often advocated as good models of the human visual system. However, there are currently many shortcomings of DCNNs, which preclude them as a model of human vision. There are continuous attempts to use features of the human visual system to improve the robustness of neural networks to data perturbations. 
We provide a detailed analysis of such bio-inspired models and their properties. To this end, we benchmark the robustness of several bio-inspired models against their most comparable baseline DCNN models. We find that bio-inspired models tend to be adversarially robust without requiring any special data augmentation. Additionally, we find that bio-inspired models beat adversarially trained models in the presence of more real-world common corruptions. Interestingly, we also find that bio-inspired models tend to use both low and mid-frequency information, in contrast to other DCNN models. We find that this mix of frequency information makes them robust to both adversarial perturbations and common corruptions.\let\thefootnote\relax\footnotetext{* The authors contributed equally.}

\end{abstract}

\section{Introduction}
\label{sec:intro}
Humans have always been the benchmark for the performance of most machine learning tasks. As deep neural networks have become more successful with several NLP and Computer Vision tasks, they get closer to beating this benchmark. However, despite their success, neural networks have several limitations in terms of distribution shifts\cite{common_corruptions}, adversarial perturbations \cite{apgd}, crowding \cite{crowding} etc. 
In particular, neural networks are extremely vulnerable to adversarial perturbations, which are small, carefully calculated perturbations that lead to gross misclassification. Since these perturbations are usually imperceptible to humans, the failure of neural networks in this area is an imminent issue. The most popular way to beat adversarial attacks is to train the network with these perturbations so that the network can learn better and avoid misclassification. This kind of data augmentation is called adversarial training and is computationally extremely expensive\cite{madry_bugs}. Furthermore, all adversarial defenses, including adversarial training, suffer from an inherent trade-off between robustness and simple classification tasks\cite{madry_bugs, hold_me_tight, preetum}. 
However, humans do not have any such trade-off and excel in most computer vision tasks. Additionally, humans do not require any additional data to be robust to perturbations. Elsayed et al.(2018) \cite{time_limited_guys} have shown that some components of the human vision could be responsible for the inherent robustness to these adversarial perturbations. Through our work, we investigate such bio-inspired models, which incorporate different components from human vision in DCNNs. We benchmark the robustness of bio-inspired models against adversarial attacks and more real-world common corruptions. Here, we find that bio-inspired models without any data augmentation tend to be surprisingly robust to small amounts of noise and have the best performance on common corruptions\cite{common_corruptions}. We also find that bio-inspired models tend to use a mix of low and mid-frequency band information, making it possible to classify more complex corruptions and be adversarially robust. Furthermore, of the the bio-inspired models analyzed, we find feedback is essential in ensuring good performance on common corruptions, showing future scope to be added to larger models.
\footnote{Code: \url{https://github.com/code-Assasin/BIIR}}
\vspace{-4pt}
\section{Bio-inspired Models}
\label{Sec:Bio_models}
\vspace{-1pt}
We have chosen four recent, publicly accessible biologically inspired models. The design of these models is explained as follows.

\textbf{Retina:} Elsayed et al.(2018) \cite{time_limited_guys} showed that the pre-processing introduced by the retina is crucial to the robustness of humans against adversarial examples. The retina introduces eccentricity-dependent sampling of the image, which creates higher resolution at the point of focus on the image and blurring in the periphery, which helps denoise the image. Additionally, this sampling happens across various focal points of the image, creating multiple viewpoint samples of the same image for better representation. Reddy et al.(2020) \cite{retina_guys} apply this non-uniform sampling as a pre-processing step for DCNNs. 

\textbf{CNNF:} Throughout the human visual system, there are multiple feed-forward and feedback connections. These feedback connections help higher-level features (like objects) modulate the lower-level feature representations (like edges) etc. This encourages the usage of representations that are more beneficial for object recognition. In fact, Elsayed et al.(2018) \cite{time_limited_guys} find that human subjects who were not given sufficient time for feedback processes to kick in were more prone to adversarial attacks compared to the ones who did not have such constraints. 
Inspired by this, Huang et al.(2020)\cite{feedback_guys} implement a feedback model wherein the input image is first pre-processed using a few convolution layers to produce a representation, $\hat{h}$. This representation goes through the feed-forward pathway to produce an output, $y$, and an intermediate representation $z$. In the feedback pass, the output $y$ is used to reconstruct both $z$ and $\hat{h}$. All the variables $\hat{h}, z $ and $y$ keep dynamically modulating each other throughout the feedback and feed-forward processes. The model learns with both standard classification and intermediate layer reconstruction losses. This gives the model the capacity to self-correct through repeated iterations of feedback and reconstruction. This is especially helpful when the images are partially occluded since the reconstruction through feedback cleans up the image for better recognition.

\textbf{MiceV1:} The inductive bias of neural networks directly impacts their performance on several tasks like generalization and robustness \cite{hold_me_tight}. Li et al., 2019 \cite{mouse_guys} hypothesize that the human visual system might have its own implicit bias, which makes it excellent at most visual tasks. If DCNNs share a similar bias, it should improve their performance. Since the bias of the human visual system is not yet quantified, the authors instead regularize the DCNN to have similar activations as those of the neural activations obtained from the V1 cortex in mice.  

\textbf{VOneNet:} Dapello et al., 2020 \cite{dicarlo} find a strong correlation between DCNNs ability to explain V1 neural responses and its adversarial robustness. This could be explained by the fact that adversarially robust models tend to use features like global information and shape \cite{madry_bugs} which is also primarily found in V1 of the human visual cortex. To optimize the usage of these V1-based features for robustness, the authors create a V1-based pre-processing block for DCNNs called VOneNet.
Its primary components include a Gabor filter bank which provides a more varied set of spatial frequency and orientation filters than traditional DCNNs. It also has a component to add stochasticity, which helps in training the network to create smoother boundaries, which prevents sudden changes in network output when small input changes are made \cite{noise_guys}. The authors train a standard DCNN with their pre-processing VOneBlock end to end and find their model to be robust to small amounts of adversarial noise. 

\vspace{-4pt}
\section{Robustness Evaluation Experiments}
\vspace{-3pt}
\label{Sec:Eval}
In order to evaluate the robustness of bio-inspired models, we firstly evaluate their adversarial robustness to $L_{\infty}$ norm-based perturbations in Sec.\ref{Sec:adv_rob}. Then, in Sec.\ref{Sec:CC} we evaluate robustness of bio-inspired models to commonly observed corruptions\cite{common_corruptions}. Lastly, in Sec.\ref{Sec:Freq_Analysis}, we explore the frequency space of these bio-inspired models to analyze the type of features they rely on, in order to explain their robustness.
We evaluate each bio-inspired model along with their standard trained baseline models and the adversarially trained models ($L_{\infty}$ norm with epsilon values of 4/255 and 8/255). For instance, for CNNF, the base architecture is a WideResNet-40-2 which when trained with standard training is labeled Baseline, and when adversarially trained with $\epsilon_{\infty}=\tfrac{4}{255}$ is labeled AT (4) and similarly for $\epsilon_{\infty}=\tfrac{8}{255}$, AT(8). We use similar nomenclature for all the bio-inspired models and their respective baseline models.
All analysis is performed on the CIFAR-10 dataset except VOneNet, which is tested on ImageNet-Val.

\subsection{Adversarial Robustness}
\begin{figure}
    \centering
    \includegraphics[width=0.95\linewidth]{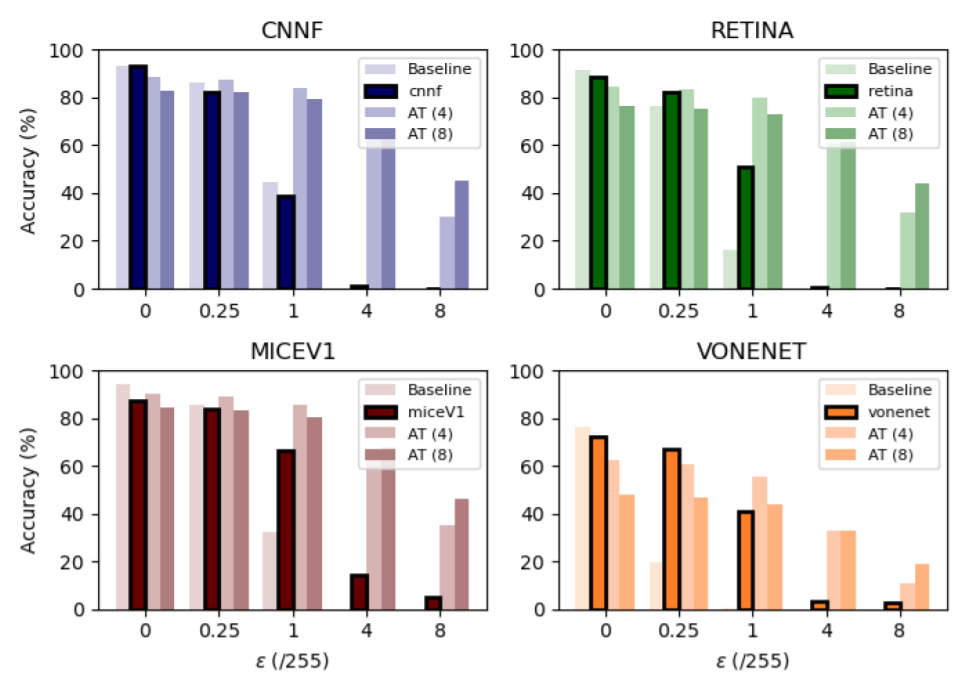}
    \caption{Robust accuracy of models against APGD attack. For all the plots, the bars with darkest colors correspond to the bio-inspired models.}
    \label{fig:apgd}
\end{figure}

\vspace{-2pt}
\label{Sec:adv_rob}
 While some of the authors of these bio-inspired models have already evaluated them, not all follow the standard evaluation protocols. Hence, we re-evaluate them on $L_{\infty}$ norm with epsilon values of $\{0, 0.25, 1, 4, 8\}$ (w.r.t 255) using the APGD attack\cite{apgd}, a powerful white-box attack. Additionally, in order to ensure fair evaluation, we follow Athalye et al.(2018)\cite{obfuscated} to break many of the bio-inspired models. We obtain at least a $3\%$ decrease in robust accuracy from the results reported by the authors (especially for CNNF).
As shown in Fig.\ref{fig:apgd}, for small epsilon values, we see that while the performance of the standard trained model drastically reduces, the performance of the bio-inspired models remains comparable or, in some cases, is even better than that of their respective adversarially trained baseline models. Additionally, unlike the adversarially trained baseline models, bio-inspired models do not have a significant trade-off for their clean accuracy ($\epsilon =0 $). For large epsilon values  ($\epsilon = 4, 8 $),  bio-inspired models fail to have a similar impact as their adversarially trained baselines. \\
However, given that both adversarially trained baseline models and the bio-inspired models do well for small epsilon values, they should be able to generalize well to more real-world corruptions \cite{maksym_cc}, which we explore in the next section.
 
\subsection{Common Corruptions}
\vspace{-2pt}
\label{Sec:CC}
 $L_{\infty}$ norm-based perturbations do not cover the kind of perturbations seen in real life. Further, humans are not only robust to $L_{\infty}$ norm perturbations\cite{time_limited_guys}, but they also do not have problems with more real-world based perturbations such as different weather conditions, noise, blurs, etc. Adversarially trained models suffer to generalize to these kinds of corruptions\cite{maksym_cc}. We test the bio-inspired models to see if the components from human vision induce sufficient robustness to such corruptions.
As shown in Table \ref{tab:cc}, on average, the Retina, VOneNet, and CNNF models beat their respective baselines. As explained in the previous section, their performance on small epsilon values indeed seems to be correlated to their performance on these corruptions.
It has been shown that adversarially trained models tend to use more features like shape and low frequencies to make decisions \cite{hold_me_tight, madry_bugs}. As shown in the Table.\ref{tab:cc}, adversarially trained models do seem to perform well for higher frequency-based corruptions (Noise corruptions). However, given that they have better shape recognition, they surprisingly fail on corruptions like fog and contrast. Through extensive analysis, Yin et al.(2019)\cite{freq_guys} also show the same, and they further explain that this is because fog and contrast-based corruptions are more complex than the others. However, as shown in Table\ref{tab:cc}, bio-inspired models have excellent performance with both fog and contrast. In particular, CNNF, due to its reconstruction and self-correcting ability (Sec.\ref{Sec:Bio_models}), performs very well on occluded images like in the case of fog.
\vspace{-5pt}
\subsection{Frequency Analysis}
\begin{figure}
    \centering
    \includegraphics[scale=0.5, width=\linewidth]{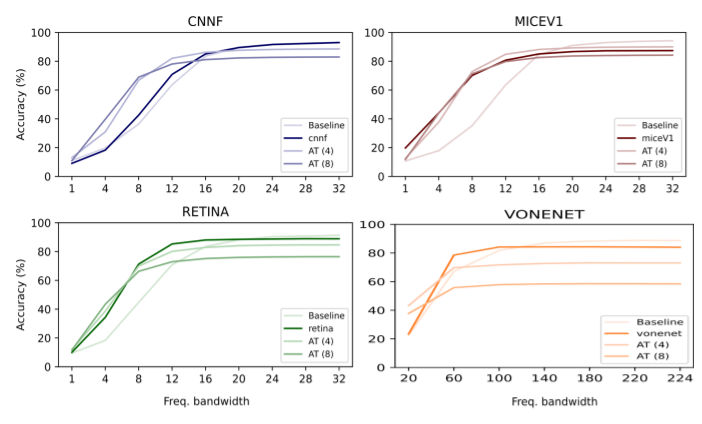}
    \caption{Accuracy of models for low frequency filtered datasets. For all the plots, the darkest colors correspond to the bio-inspired models.}
    \label{fig:low_freq}
\end{figure}

\vspace{-1pt}
\label{Sec:Freq_Analysis}
As we have seen, despite the use of global information like shape, adversarially trained models have a hard time with heavily occluded corruptions like Fog (Tab.\ref{tab:cc}). We probe to see what makes bio-inspired models unique against such corruptions. 
To examine this, we create low-pass filtered datasets to analyze the frequency information that bio-inspired models use. We create these datasets by applying a low pass filter of a specific bandwidth to all the images in the dataset \cite{hold_me_tight}. As shown in Fig.\ref{fig:low_freq}, bio-inspired models indeed use low-frequency information but not as well as their adversarially trained baseline models, which also explains the robustness of the former to small amounts of adversarial noise \cite{madry_bugs}. However, as more mid-frequency information is added (bandwidth 8-20 for Cifar; 40-100 for Imagenet \cite{freq_guys,vitcnn}), bio-inspired models see the largest surge in accuracy. When higher frequency information is added, the standard trained baseline models see more increase in accuracy while the other models do not gain much.
To summarize, bio-inspired models tend to rely on a mix of low and mid-frequency band information compared to their baseline models. This mix of frequency information explains why they can handle more complex corruptions like fog and contrast very well. As shown by Yin et al.(2019)\cite{freq_guys} an over-reliance on low-frequency information worsens the performance on these more complex corruptions, which is why adversarially trained models perform extremely poorly on them. However, bio-inspired models do not face any trouble with the same. 

\section{Conclusion and Future Work}
Our analysis shows that bio-inspired models present a low-cost alternative to adversarial training. They are robust to small amounts of adversarial noise and do extremely well for real-world corruptions. For larger adversarial perturbations, we hypothesize that bio-inspired models may need to utilize a stronger base architecture. It has been shown that the trade-off between adversarial robustness and classification may be due to an inherent lack of capacity of our current DCNN models \cite{preetum}. Bio-inspired models do present simple ways to increase the capacity of these DCNNs and indeed reduce the trade-off encountered. 
In particular, feedback pathways (CNNF) provide an efficient way of increasing the capacity of models atleast by a factor of the number of self-iterations\cite{serre_feedback}. We have seen that feedback pathways play a large role in the model's robustness. Hence, as part of our future work, we would like to explore the use of such components in stronger, higher capacity architectures like Vision Transformers\cite{vitcnn}.

\vspace{-5pt}

\begin{table*}[]
\centering
\resizebox{18cm}{!} {
\begin{tabular}{@{}|c|ccc|cccc|cccc|cccc|c|@{}}
\toprule
Corruptions  ($\xrightarrow{}$) & \multicolumn{3}{c|}{Noise} & \multicolumn{4}{c|}{Blur} & \multicolumn{4}{c|}{Weather} & \multicolumn{4}{c|}{Digital} &  \\ \cmidrule(r){1-16}
Models ($\downarrow$) & \multicolumn{1}{c|}{\begin{tabular}[b]{@{}c@{}}Gaussian\\ Noise\end{tabular}} & \multicolumn{1}{c|}{\begin{tabular}[b]{@{}c@{}}Shot \\ Noise\end{tabular}} & \begin{tabular}[b]{@{}c@{}}Impulse\\ Noise\end{tabular} & \multicolumn{1}{c|}{\begin{tabular}[b]{@{}c@{}}Defocus \\ Blur\end{tabular}} & \multicolumn{1}{c|}{\begin{tabular}[b]{@{}c@{}}Glass \\ Blur\end{tabular}} & \multicolumn{1}{c|}{\begin{tabular}[b]{@{}c@{}}Motion \\ Blur\end{tabular}} & \begin{tabular}[b]{@{}c@{}}Zoom \\ Blur\end{tabular} & \multicolumn{1}{c|}{Frost} & \multicolumn{1}{c|}{Snow} & \multicolumn{1}{c|}{Fog} & \begin{tabular}[b]{@{}c@{}}Bright \\ness.\end{tabular} & \multicolumn{1}{c|}{\begin{tabular}[b]{@{}c@{}}Cont \\ -rast.\end{tabular}} & \multicolumn{1}{c|}{\begin{tabular}[b]{@{}c@{}}Elastic \\ Trans.\end{tabular}} & \multicolumn{1}{c|}{Pixelate} & \begin{tabular}[b]{@{}c@{}}JPEG \\ Comp.\end{tabular} & Mean\\
\midrule
Retina Baseline & \multicolumn{1}{c|}{39.51} & \multicolumn{1}{c|}{50.95} & 52.14 & \multicolumn{1}{c|}{77.52} & \multicolumn{1}{c|}{44.58} & \multicolumn{1}{c|}{66.97} & 70.34 & \multicolumn{1}{c|}{70.15} & \multicolumn{1}{c|}{75.03} & \multicolumn{1}{c|}{81.37} & \textbf{89.3} & \multicolumn{1}{c|}{66.5} & \multicolumn{1}{c|}{77.29} & \multicolumn{1}{c|}{66.99} & 74.89 & 66.9 \\
Retina AT (4) & \multicolumn{1}{c|}{\textbf{79.42}} & \multicolumn{1}{c|}{\textbf{80.65}} & \textbf{74.7} & \multicolumn{1}{c|}{79.64} & \multicolumn{1}{c|}{\textbf{77.08}} & \multicolumn{1}{c|}{75.41} & 78.87 & \multicolumn{1}{c|}{76.2} & \multicolumn{1}{c|}{\textbf{78.58}} & \multicolumn{1}{c|}{59.97} & 82.22 & \multicolumn{1}{c|}{44.52} & \multicolumn{1}{c|}{78.85} & \multicolumn{1}{c|}{82.5} & \textbf{82.57} & 75.41 \\
Retina AT (8) & \multicolumn{1}{c|}{73.3} & \multicolumn{1}{c|}{74.2} & 71.87 & \multicolumn{1}{c|}{72.45} & \multicolumn{1}{c|}{70.77} & \multicolumn{1}{c|}{68.81} & 71.76 & \multicolumn{1}{c|}{64.7} & \multicolumn{1}{c|}{70.8} & \multicolumn{1}{c|}{52.02} & 72.31 & \multicolumn{1}{c|}{38.46} & \multicolumn{1}{c|}{71.15} & \multicolumn{1}{c|}{74.67} & 74.78 & 68.14 \\
Retina & \multicolumn{1}{c|}{46.83} & \multicolumn{1}{c|}{55.02} & 56.74 & \multicolumn{1}{c|}{87.74} & \multicolumn{1}{c|}{70.19} & \multicolumn{1}{c|}{\textbf{83.69}} & \textbf{87.79} & \multicolumn{1}{c|}{\textbf{76.66}} & \multicolumn{1}{c|}{77.93} & \multicolumn{1}{c|}{\textbf{80.6}} & 87 & \multicolumn{1}{c|}{\textbf{68.12}} & \multicolumn{1}{c|}{\textbf{86.33}} & \multicolumn{1}{c|}{\textbf{86.01}} & 81.43 & \textbf{75.47} \\ \midrule
MiceV1 Baseline & \multicolumn{1}{c|}{41.3} & \multicolumn{1}{c|}{54.91} & 49.49 & \multicolumn{1}{c|}{82.29} & \multicolumn{1}{c|}{48.92} & \multicolumn{1}{c|}{77.19} & 76.8 & \multicolumn{1}{c|}{78.99} & \multicolumn{1}{c|}{81.56} & \multicolumn{1}{c|}{\textbf{88.28}} & \textbf{92.91} & \multicolumn{1}{c|}{\textbf{75.15}} & \multicolumn{1}{c|}{\textbf{82.99}} & \multicolumn{1}{c|}{75.61} & 77.79 & 72.28 \\
MiceV1 AT (4) & \multicolumn{1}{c|}{\textbf{84.03}} & \multicolumn{1}{c|}{\textbf{85.33 }} & \textbf{75.8} & \multicolumn{1}{c|}{\textbf{84.38}} & \multicolumn{1}{c|}{\textbf{81.52}} & \multicolumn{1}{c|}{\textbf{79.83}} & \textbf{83.43} & \multicolumn{1}{c|}{\textbf{81.46}} & \multicolumn{1}{c|}{\textbf{83.19}} & \multicolumn{1}{c|}{64.94} & 86.93 & \multicolumn{1}{c|}{49.57} & \multicolumn{1}{c|}{83.7} & \multicolumn{1}{c|}{\textbf{87.72}} & \textbf{87.4} & \textbf{79.95} \\
MiceV1 AT (8) & \multicolumn{1}{c|}{79.67} & \multicolumn{1}{c|}{80.55} & 74.73 & \multicolumn{1}{c|}{79.35} & \multicolumn{1}{c|}{77.87} & \multicolumn{1}{c|}{75.45} & 78.19 & \multicolumn{1}{c|}{73.89} & \multicolumn{1}{c|}{78.55} & \multicolumn{1}{c|}{58.56} & 80.46 & \multicolumn{1}{c|}{42.81} & \multicolumn{1}{c|}{78.36} & \multicolumn{1}{c|}{81.95} & 82 & 74.83 \\
MiceV1 & \multicolumn{1}{c|}{66.35} & \multicolumn{1}{c|}{70.59} & 63.46 & \multicolumn{1}{c|}{71.84} & \multicolumn{1}{c|}{71.9} & \multicolumn{1}{c|}{61.45} & 66.81 & \multicolumn{1}{c|}{73.79} & \multicolumn{1}{c|}{70.36} & \multicolumn{1}{c|}{71.55} & 81.52 & \multicolumn{1}{c|}{62.1} & \multicolumn{1}{c|}{73.34} & \multicolumn{1}{c|}{73.92} & 80.64 & 70.64 \\ \midrule
CNNF Baseline & \multicolumn{1}{c|}{42.77} & \multicolumn{1}{c|}{54.45} & 44.49 & \multicolumn{1}{c|}{77.23} & \multicolumn{1}{c|}{44.35} & \multicolumn{1}{c|}{69.52} & 70.23 & \multicolumn{1}{c|}{72.3} & \multicolumn{1}{c|}{78.65} & \multicolumn{1}{c|}{80.06} & 91.31 & \multicolumn{1}{c|}{59.85} & \multicolumn{1}{c|}{81.01} & \multicolumn{1}{c|}{70.95} & 75.86 & 67.53 \\
CNNF AT (4) & \multicolumn{1}{c|}{\textbf{81.36}} & \multicolumn{1}{c|}{\textbf{83.35}} & 67.5 & \multicolumn{1}{c|}{80.36} & \multicolumn{1}{c|}{\textbf{78.72}} & \multicolumn{1}{c|}{74.26} & 78.72 & \multicolumn{1}{c|}{\textbf{79.27}} & \multicolumn{1}{c|}{\textbf{82.42}} & \multicolumn{1}{c|}{62.33} & 85.89 & \multicolumn{1}{c|}{47.4} & \multicolumn{1}{c|}{80} & \multicolumn{1}{c|}{\textbf{86.05}} & \textbf{85.48} & 76.87 \\
CNNF AT (8) & \multicolumn{1}{c|}{79.67} & \multicolumn{1}{c|}{80.59} & \textbf{75.91} & \multicolumn{1}{c|}{76.9} & \multicolumn{1}{c|}{76.15} & \multicolumn{1}{c|}{72.32} & 75.58 & \multicolumn{1}{c|}{71.38} & \multicolumn{1}{c|}{78.01} & \multicolumn{1}{c|}{56.83} & 79.1 & \multicolumn{1}{c|}{40.44} & \multicolumn{1}{c|}{76.02} & \multicolumn{1}{c|}{80.62} & 80.87 & 73.36 \\

CNNF & \multicolumn{1}{c|}{56.74} & \multicolumn{1}{c|}{65.85} & 63.36 & \multicolumn{1}{c|}{\textbf{86.35}} & \multicolumn{1}{c|}{51.03} & \multicolumn{1}{c|}{\textbf{82.3}} & \textbf{83.61} & \multicolumn{1}{c|}{78.92} & \multicolumn{1}{c|}{81.31} & \multicolumn{1}{c|}{\textbf{89.51}} & \textbf{91.7} & \multicolumn{1}{c|}{\textbf{90.13}} & \multicolumn{1}{c|}{\textbf{83.63}} & \multicolumn{1}{c|}{74.51} & 79.04 & \textbf{77.93} \\ \midrule
Vonenet Baseline & \multicolumn{1}{c|}{31.12} & \multicolumn{1}{c|}{28.6} & 26.59 & \multicolumn{1}{c|}{\textbf{35.52}} & \multicolumn{1}{c|}{25.48} & \multicolumn{1}{c|}{\textbf{36.35}} & \textbf{36.32} & \multicolumn{1}{c|}{\textbf{35.17}} & \multicolumn{1}{c|}{\textbf{30.54}} & \multicolumn{1}{c|}{\textbf{43.41}} & \textbf{65.19} & \multicolumn{1}{c|}{\textbf{35.95}} & \multicolumn{1}{c|}{43.34} & \multicolumn{1}{c|}{45.55} & 52.44 & 38.11 \\
Vonenet AT (4) & \multicolumn{1}{c|}{27.95} & \multicolumn{1}{c|}{26.64} & 20.42 & \multicolumn{1}{c|}{21.8} & \multicolumn{1}{c|}{30.28} & \multicolumn{1}{c|}{29.44} & 32.36 & \multicolumn{1}{c|}{29.77} & \multicolumn{1}{c|}{28.23} & \multicolumn{1}{c|}{7.01} & 50.82 & \multicolumn{1}{c|}{8.36} & \multicolumn{1}{c|}{45.25} & \multicolumn{1}{c|}{52.66} & 55.21 & 31.08 \\
Vonenet AT (8) & \multicolumn{1}{c|}{24.26} & \multicolumn{1}{c|}{22.67} & 17.44 & \multicolumn{1}{c|}{16.73} & \multicolumn{1}{c|}{23.15} & \multicolumn{1}{c|}{22.06} & 24.23 & \multicolumn{1}{c|}{17.34} & \multicolumn{1}{c|}{21.64} & \multicolumn{1}{c|}{2.02} & 35.46 & \multicolumn{1}{c|}{2.89} & \multicolumn{1}{c|}{35.64} & \multicolumn{1}{c|}{39.7} & 43.08 & 23.22 \\
Vonenet & \multicolumn{1}{c|}{\textbf{33.18}} & \multicolumn{1}{c|}{\textbf{31.73}} & \textbf{30.45} & \multicolumn{1}{c|}{34.34} & \multicolumn{1}{c|}{\textbf{32.85}} & \multicolumn{1}{c|}{34.68} & 34.33 & \multicolumn{1}{c|}{35.3} & \multicolumn{1}{c|}{25.27} & \multicolumn{1}{c|}{30.38} & 60.93 & \multicolumn{1}{c|}{27.5} & \multicolumn{1}{c|}{\textbf{46.71}} & \multicolumn{1}{c|}{\textbf{58.37}} & \textbf{57.04} & \textbf{38.2} \\ \bottomrule
\end{tabular}
}
\caption{Common Corruptions accuracies across different models. We can see the bad performance of adversarially trained models on weather-based corruptions. For VOneNet, since it is tested on a larger dataset (ImageNet-C), the performance improvement is not as drastic as CNNF (Sec.\ref{Sec:CC}).   }

\label{tab:cc} 
\end{table*}


{\small
\bibliographystyle{ieee_fullname}
\bibliography{egbib}
}

\end{document}